\documentclass[letterpaper, 10 pt, conference]{ieeeconf}  

\IEEEoverridecommandlockouts                              

\usepackage{graphicx} 
\usepackage{amsmath} 
\usepackage{amssymb}  
\usepackage[table,xcdraw]{xcolor}
\usepackage{subfigure}
\usepackage{tabularx}
\usepackage{cite}

\makeatletter
\let\NAT@parse\undefined
\makeatother
\usepackage[numbers]{natbib}
\usepackage{siunitx}
\usepackage{makecell} 

\usepackage{aprl_acronyms}
\usepackage{aprl_misc}
\usepackage{multirow}

\let\labelindent\relax
\usepackage{enumitem}

\usepackage{soul,color}

\title{
\LARGE \bf Have We Mastered Scale in Deep Monocular Visual SLAM? \\ The ScaleMaster Dataset and Benchmark
}

\author{Hyoseok Ju$^{1}$, Bokeon Suh$^{1}$, and Giseop Kim$^{1*}$%
\thanks{$^{1}$H. Ju, B. Suh, and G. Kim are with the Department of Robotics and Mechatronics Engineering, DGIST, Daegu, Republic of Korea {\tt\small [hyoseok.ju, bokeon.suh, gsk]@dgist.ac.kr}}%
}

\usepackage[colorlinks=true, citecolor=blue, linkcolor=cyan, urlcolor=magenta]{hyperref}
\usepackage[font=small,labelfont=bf,labelsep=period]{caption}
\usepackage{threeparttable}
\usepackage{svg}
\usepackage[percent]{overpic}
\usepackage{dblfloatfix}
\usepackage{booktabs}
\usepackage{array}
\usepackage{xcolor}

\begin{document}

\maketitle
\thispagestyle{empty}
\pagestyle{empty}


\begin{abstract}
Recent advances in deep monocular visual \ac{SLAM} have achieved impressive accuracy and dense reconstruction capabilities, yet their robustness to scale inconsistency in large-scale indoor environments remains largely unexplored. Existing benchmarks are limited to room-scale or structurally simple settings, leaving critical issues of intra-session scale drift and inter-session scale ambiguity insufficiently addressed. To fill this gap, we introduce the \textit{ScaleMaster Dataset}, the first benchmark explicitly designed to evaluate scale consistency under challenging scenarios such as multi-floor structures, long trajectories, repetitive views, and low-texture regions. We systematically analyze the vulnerability of state-of-the-art deep monocular visual SLAM systems to scale inconsistency, providing both quantitative and qualitative evaluations. Crucially, our analysis extends beyond traditional trajectory metrics to include a direct map-to-map quality assessment using metrics like Chamfer distance against high-fidelity 3D ground truth. Our results reveal that while recent deep monocular visual \ac{SLAM} systems demonstrate strong performance on existing benchmarks, they suffer from severe scale-related failures in realistic, large-scale indoor environments. By releasing the ScaleMaster dataset and baseline results, we aim to establish a foundation for future research toward developing scale-consistent and reliable visual \ac{SLAM} systems.

\end{abstract}

\section{Introduction} 
\label{sec:intro}

Visual \ac{SLAM} \cite{cadena2017past,ORBSLAM3_TRO,slam-handbook} is experiencing a paradigm shift toward dense mapping methods that generate accurate 3D maps in real time. Beyond traditional feature-based pipelines \cite{ORBSLAM3_TRO}, recent approaches leverage feed-forward dense pointmap estimators \cite{dust3r_cvpr24} as a front-end to directly reconstruct scenes while estimating camera poses. These methods achieve both visually detailed reconstructions and high-precision trajectory estimation. A representative example is MASt3R-SLAM \cite{murai2025mast3r}, which employs the direct pointmap estimator \cite{dust3r_cvpr24,leroy2024grounding} to generate detailed 3D maps and has demonstrated state-of-the-art performance on standard indoor benchmarks.

\begin{figure}[!t]
    \centering
    \includegraphics[width=0.98\columnwidth]{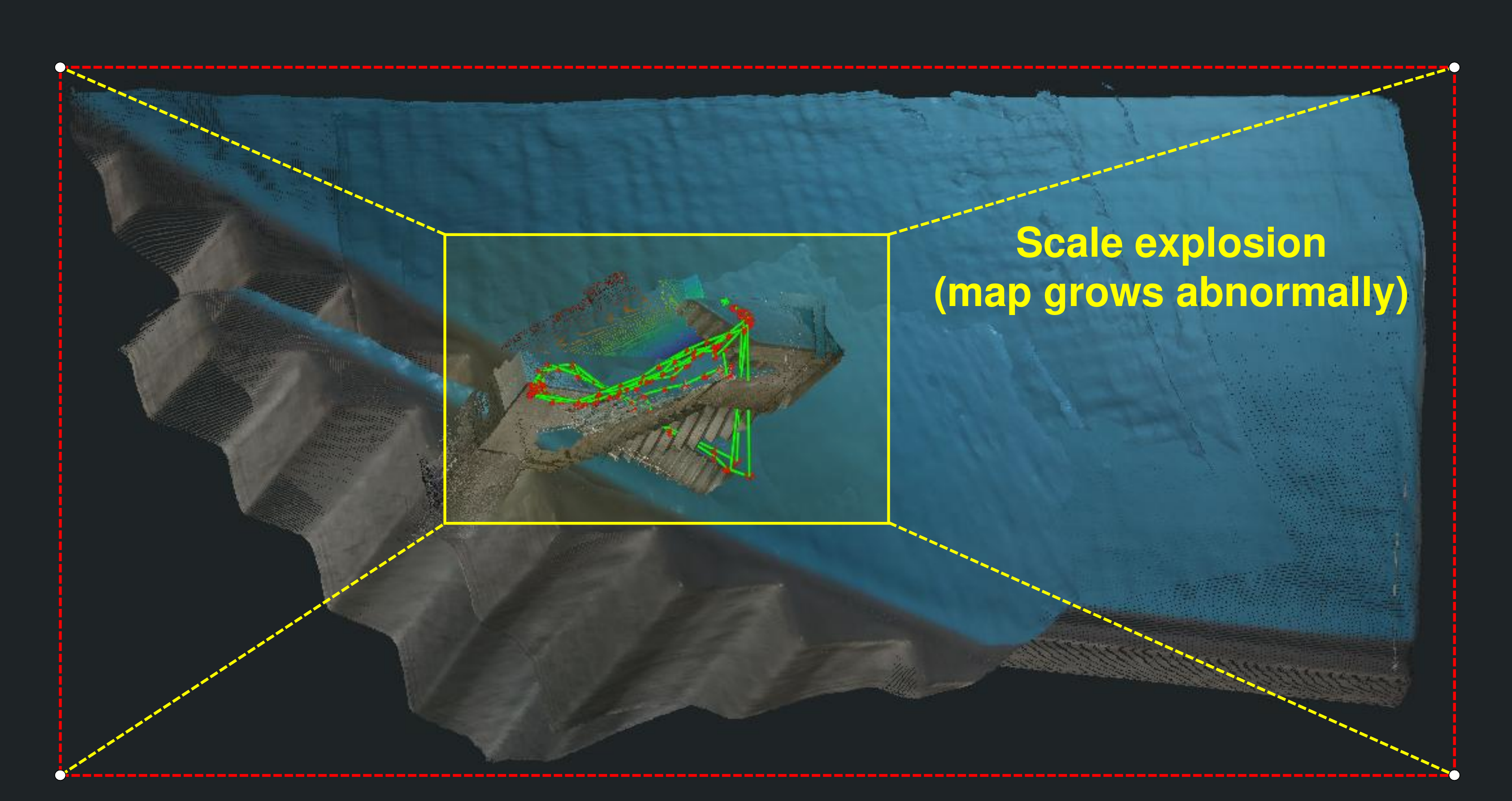}
    \caption{Representative example of scale inconsistency. During Sim(3) pose-graph optimization, the SLAM trajectory experiences a sudden scale explosion (highlighted), resulting in an abnormally enlarged reconstruction that diverges from the previously estimated map.}

    \vspace{-3mm}
    
    \label{fig:representative_figure}
\end{figure}

However, the outstanding performance of these modern dense visual SLAM systems has been predominantly validated on benchmarks featuring either room-scale environments such as TUM-RGBD \cite{sturm2012benchmark}, or structurally simple spaces like those in EuRoC \cite{burri2016euroc}. Consequently, their robustness in large-scale, complex indoor environments (e.g., multi-floor structures, vertical motions, long loops) remains largely unverified. Specifically, the critical issues of \textbf{scale ambiguity} and \textbf{scale inconsistency} that arise over long trajectories have not been sufficiently addressed as shown in \figref{fig:representative_figure}. Furthermore, evaluation has traditionally focused on trajectory error (e.g., \ac{ATE}), a metric that we will demonstrate can be insufficient and even misleading. Even when poses look accurate after Sim(3) pose-graph optimization, the dense map may still be inconsistent without a joint Sim(3) bundle adjustment, as observed in systems like MASt3R-SLAM. Thus, a direct map-to-map comparison provides a more faithful way to reveal hidden inconsistencies such as warping and geometric distortion that trajectory-only metrics tend to overlook.

This paper aims to provide an in-depth analysis of the scale inconsistency problem that arises in large-scale, complex indoor environments—involving trajectories of several hundred meters and multi-floor structures—which have not been deeply addressed by existing benchmarks (e.g., \cite{sturm2012benchmark,burri2016euroc,glocker2013real}). To this end, we introduce the \textbf{ScaleMaster} dataset\footnote{https://scalemaster-dataset.github.io/}, the first benchmark intentionally designed to target scale inconsistency by incorporating challenging scenarios where this problem becomes prominent. The ScaleMaster dataset provides a systematic empirical environment to evaluate and analyze scale inconsistency in depth, thereby overcoming the limitations of existing benchmarks.

In this paper, we make the following key contributions:

\begin{itemize}

    \item \textbf{Systematic Analysis of Scale Inconsistencies:} We analyze how state-of-the-art monocular deep visual SLAM systems encounter \textbf{scale inconsistency} issues (as taxonomically analyzed in \figref{fig:problem_diagram}) in large-scale and complex indoor environments, highlighting vulnerabilities that are not fully revealed by existing benchmarks.
    
    \item \textbf{ScaleMaster Dataset and Baseline Evaluation:} We introduce the ScaleMaster Dataset, designed to examine scale consistency in challenging settings such as multi-floor structures, long trajectories, repetitive patterns, and low-texture areas. Using this dataset, we conduct baseline evaluations of representative deep-learning-based SLAM systems (DROID-SLAM \cite{teed2021droid}, MASt3R-SLAM \cite{murai2025mast3r}, VGGT-SLAM \cite{maggio2025vggt}), and reveal concrete cases of intra-session scale drift and inter-session scale ambiguity through both quantitative and qualitative analysis.
    
    \item \textbf{Complementary Use of Map Quality Metrics:} To address limitations of trajectory-only evaluation, we incorporate map-to-map metrics (Chamfer distance and Drop Rate), showing how they complement ATE by revealing distortions and scale collapse that remain hidden otherwise.

\end{itemize}

\section{Related Works}
\label{sec:rel}

\subsection{Deep Learning-based Visual SLAM}
\label{sec:rel1}

Following traditional modular visual SLAM systems \cite{murTRO2015,klein2007parallel}, research integrating deep learning into the SLAM pipeline has become mainstream \cite{favorskaya2023deep,teed2021droid,zhu2022nice}. A prominent example of this trend is \text{DROID-SLAM} \cite{teed2021droid}, which set a new standard by achieving high accuracy through a fully differentiable architecture. More recently, a subsequent wave of \text{feed-forward dense SLAM} systems has emerged, which leverage pre-trained two-view reconstruction foundation models. \text{MASt3R-SLAM} \cite{murai2025mast3r} and \text{VGGT-SLAM} \cite{maggio2025vggt}, which attempt to address the ambiguity of the scale, are notable examples of this approach. The aforementioned works aim to maintain the robustness of modern learning-based \ac{SLAM} systems in real-world environments. However, their robustness to scale drift, particularly in large-scale and complex indoor environments, remains largely unexplored.

\subsection{Benchmark Datasets for Visual SLAM}
\label{sec:rel2}
The progress and evaluation of visual SLAM algorithms have been critically dependent on the role of standard benchmark datasets that provide precise ground truth. The rigorous performance assessment in SLAM research is primarily based on three key datasets. \text{TUM-RGBD} dataset \cite{sturm2012benchmark}  established the standard for evaluating \ac{ATE}, while the \text{EuRoC MAV} dataset \cite{burri2016euroc} has served as a crucial benchmark for the robustness of visual-inertial systems. Lastly, \text{7-Scenes} \cite{glocker2013real} is specialized for evaluating re-localization performance. These are collectively regarded as the de facto standards for performance validation in the visual SLAM community. Despite their utility, standard benchmarks are mostly room-scale and do not support systematic evaluation of long-term \textbf{intra-session} scale drift or \textbf{inter-session} scale consistency. ARKitScenes \cite{dehghan2021arkitscenes}—though a comparatively recent dataset—also falls short, as it does not encompass complex indoor environments with long trajectories. This lack of a scale-focused benchmark has limited systematic analysis of scale failure modes; this is a gap we address in this work.

\subsection{Evaluation Metrics for Visual \ac{SLAM}}
\label{sec:rel3}
A commonly reported trajectory metric is the \ac{ATE} \cite{Zhang18iros}, computed after Sim(3) alignment of the estimated poses to ground truth; it summarizes global trajectory agreement as a \ac{RMSE} over pose errors. While invaluable, \ac{ATE} only assesses the accuracy of the localization component (L in \ac{SLAM}). It provides no direct information about the quality of the mapping component (M), which is the primary output of dense \ac{SLAM} systems \cite{li2023dense}. To directly assess 3D structure quality, we adopt the Chamfer distance \cite{murai2025mast3r} (and Drop Rate will be defined in \secref{sec:exp-setup}) between the reconstructed map and a high-fidelity reference point cloud. These map-oriented metrics provide geometry-focused signals that complement trajectory-only evaluation and have been used in prior dense visual SLAM work. Although many \ac{SLAM} datasets exist (e.g., \cite{wei2025fusionportablev2,chen2024heterogeneous}), the benchmarks that deep, dense monocular visual SLAM systems have been primarily evaluated on are those highlighted in \secref{sec:rel2} (e.g., \cite{sturm2012benchmark,burri2016euroc,glocker2013real}). These room-scale or structurally simple indoor datasets do not capture challenging large-volume environments (e.g., multi-floor lobbies with tens of meters of ceiling height), where long-range intra- and inter-session scale inconsistency becomes apparent. Because dense visual SLAM yields dense maps as well as poses, map-oriented evaluation is crucial: ATE can look correct while the map suffers scale distortions. With Chamfer- and Drop Rate–based evaluation on ScaleMaster, we provide a protocol that exposes such failures under conditions existing benchmark datasets cannot reveal.



%
\begin{figure}[!t]
    \centering
    \includegraphics[width=0.98\columnwidth]{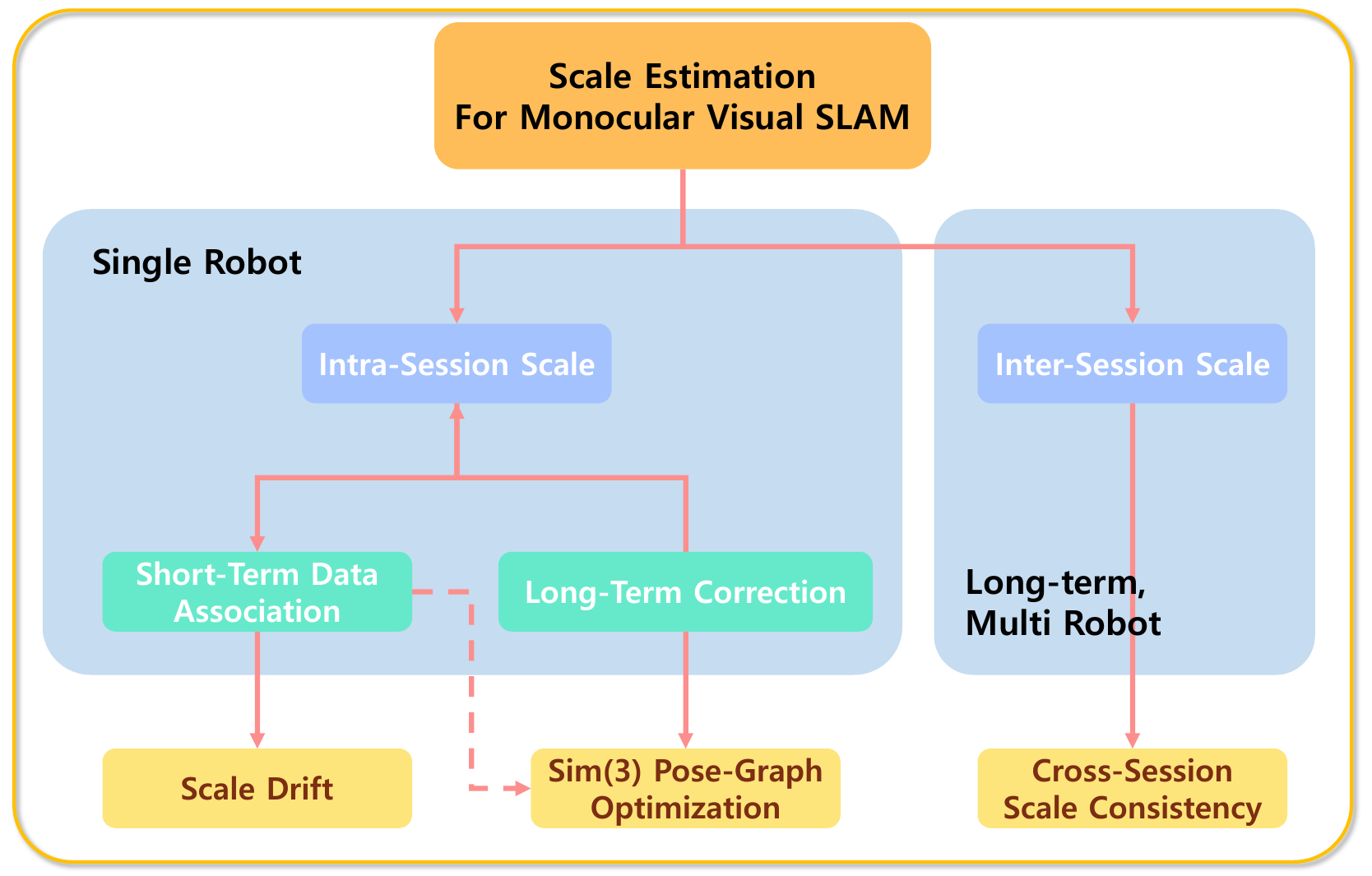}
    \caption{A diagram illustrating the components of the scale estimation problem.}

    \vspace{-3mm}
    
    \label{fig:problem_diagram}
\end{figure}
%

\section{Problem Definition}
\label{sec:prob_def}

\begin{table*}[!t]
\centering
\caption{\textbf{Comparison of public benchmark datasets with ours.}}
\resizebox{0.85\textwidth}{!}{%
\begin{tabular}{c|ccc|cccc}
\midrule
\textbf{Dataset} & \textbf{Sequences} & \textbf{\begin{tabular}[c]{@{}c@{}}Avg. \\ Length\end{tabular}} & \textbf{Image Resolution} & \textbf{\begin{tabular}[c]{@{}c@{}}Multi-floor \& \\ Elevation\end{tabular}} & \textbf{\begin{tabular}[c]{@{}c@{}}Pure \\ Rotation\end{tabular}} & \textbf{\begin{tabular}[c]{@{}c@{}}Complex \\ Indoor\end{tabular}} & \textbf{\begin{tabular}[c]{@{}c@{}}Scale Study \\ Feasibility\end{tabular}} \\ \midrule
\textbf{EuRoC} & 11 & 81.2 m & $752 \times 480$ & $\triangle$ & X & X & X \\
\textbf{TUM-RGBD} & $\sim$39 & 12.2 m & $640 \times 480$ & X & X & X & X \\
\textbf{7-Scenes} & 7 & 64.3 m & $640 \times 480$ & X & X & X & X \\
\textbf{ARKitScenes} & \textgreater 5,000 & \textless 100 m & $1920 \times 1440$ & X & O & X & X \\ \midrule
\textbf{Ours (ScaleMaster dataset)} & 25 & 152.2 m & $1920 \times 1440$ & O & O & O & O \\ \midrule
\end{tabular}
}
\label{dataset_comparison}
\end{table*}
\begin{figure*}[t]
    \centering
    \includegraphics[width=0.98\textwidth]{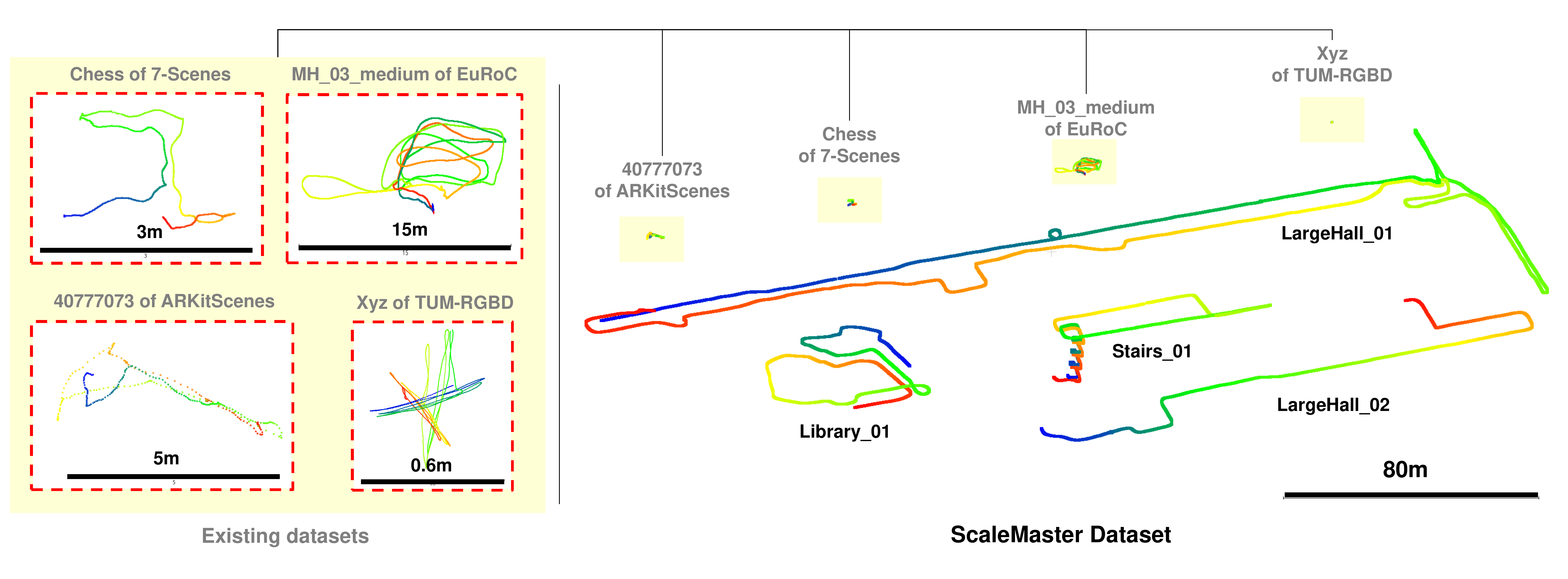}
    \caption{A visual comparison of trajectory scales between our proposed ScaleMaster dataset and existing standard benchmarks. This distinct contrast visually demonstrates why existing room-scale benchmarks are insufficient for evaluating long-term scale consistency failures.}
    \label{fig:trajectory_comparison}

    \vspace{-3mm}

\end{figure*}

The scale consistency problem in monocular visual \ac{SLAM} can be categorized into two primary challenges as in \figref{fig:problem_diagram}: Intra-session inconsistency, which occurs within a single run, and inter-session ambiguity, which arises between multiple runs (e.g., long-term map management or multi-robot collaborative mapping). 

\subsection{Intra-Session Scale Inconsistency}
Intra-session scale inconsistency refers to the scale errors that accumulate during a single \ac{SLAM} session. This problem arises primarily from the interplay of the following two factors:

\begin{itemize}    
    \item \textbf{Short-Term Data Association \& Scale Drift:} 
     Short-term data association establishes odometry edges by relating consecutive image frames. Due to the inherent unobservability of metric scale in monocular visual SLAM, relative scale drift inevitably arises between them \cite{hartley2003multiple}.

    \item \textbf{Long-Term Correction \& Limitations of $\text{Sim}(3)$ \ac{PGO}:} 
    Owing to the $\text{Sim}(3)$ gauge in monocular visual SLAM, scale cannot be fixed by $\text{SE}(3)$ optimization \cite{strasdat2010scale}; while $\text{Sim}(3)$ pose-graph optimization is standard, its small, Gaussian-error assumption makes it brittle to large, discontinuous loop-closure errors, leaving residual scale inconsistency.
    
\end{itemize}

\subsection{Inter-Session Scale Ambiguity}
\label{sec:intersessionscale}
Inter-session scale ambiguity arises when attempting to merge multiple maps generated at different times or when trying to re-localize within a map from a previous session \cite{kim2010multiple}. If each session already contains its own unique and uncorrected scale drift, determining the relative scale between these maps becomes an extremely challenging problem. This ultimately leads to the failure of \textbf{Cross-Session Scale Consistency}, making it impossible to integrate multiple maps into a single, globally consistent representation.

\renewcommand{\arraystretch}{1.2}
\begin{table*}[t]
\centering
\caption{\textbf{Summary of collected sequences and their characteristics.}}
\begin{threeparttable}
\resizebox{0.98\textwidth}{!}{%
\begin{tabular}{lccc p{5.5cm} p{4.2cm}}
\midrule
\textbf{Sequence Name} & \textbf{Frames} & \textbf{Path Length (m)} & \textbf{Duration (s)} & \textbf{Environment} & \textbf{Tags} \\ \midrule
\textbf{Basement\_01}   & 2036  & 29.11  & 67  & Basement area followed by a staircase ascent. & Indoor, Short trajectory, 3D Map \\
\textbf{HotelRoom\_01}  & 4217  & 29.07  & 141 & Interior traversal of a hotel room. & Indoor, Repetitive view, Short trajectory \\
\textbf{Lab\_01}         & 2167  & 25.85  & 72  & In-place rotations inside a lab room. & Indoor, Repetitive view, Short trajectory \\
\textbf{LargeHall\_01}  & 22830 & 884.12 & 761 & Full loop covering the entire LargeHall. & Indoor, Very long trajectory \\
\textbf{LargeHall\_02}  & 6576  & 241.89 & 219 & Evening traversal of a large open hall, low-light. & Indoor, Long trajectory, 3D Map \\
\textbf{LargeHall\_03}  & 2764  & 109.89 & 92  & Short loop inside the LargeHall at night. & Indoor, Low-texture risk, Medium trajectory \\
\textbf{LargeHall\_04}  & 4331  & 179.69 & 144 & Loop around the E1 section of LargeHall. & Indoor, Medium trajectory \\
\textbf{LargeHall\_05}  & 1912  & 54.21  & 63  & Traversal under low-light conditions. & Indoor, Short trajectory, 3D Map \\
\textbf{Library\_01}    & 6515  & 254.98 & 217 & Multi-floor descent from 5F to 3F with loops per floor. & Indoor, Long trajectory, Repetitive view, 3D Map \\
\textbf{Library\_02}    & 5001  & 163.58 & 166 & Single-floor loop on 4F with repetitive bookshelves. & Indoor, Medium trajectory, 3D Map \\
\textbf{Library\_03}    & 3136  & 105.81 & 105 & Loop on the 3rd floor of the library. & Indoor, Medium trajectory \\
\textbf{Library\_04}    & 5450  & 146.22 & 182 & Walking paths between library bookshelves. & Indoor, Repetitive view, Medium trajectory \\
\textbf{Library\_05}    & 2540  & 78.24  & 85  & Large open central atrium (depth sensor limitation). & Indoor, Low-texture risk, Short trajectory \\
\textbf{Library\_06}    & 2026  & 13.27  & 67  & 360-degree in-place rotation at the library center. & Indoor, Short trajectory, 3D Map \\
\textbf{Library\_07}    & 1580  & 5.23   & 52  & Static panoramic survey from the 1st floor. & Indoor, Short trajectory, 3D Map \\
\textbf{Library\_08}    & 2241  & 3.51   & 75  & Short traversal around the open central viewpoint. & Indoor, Low-texture risk, Short trajectory \\
\textbf{Library\_09}    & 2303  & 20.65  & 77  & In-place rotation in front of a 3F glass room. & Indoor, Repetitive view, Short trajectory \\
\textbf{Lobby\_01}      & 2893  & 104.83 & 96  & Traversal inside a lobby. & Indoor, Medium trajectory \\
\textbf{Lounge\_01}     & 6823  & 199.09 & 228 & Loop trajectory inside a lounge area. & Indoor, Medium trajectory \\
\textbf{Office\_01}     & 6009  & 154.50 & 200 & Traversal inside a repetitive office view. & Indoor, Repetitive view, Medium trajectory \\
\textbf{Parking\_01}    & 8218  & 323.01 & 274 & Full loop inside the underground parking lot (B2). & Underground, Long trajectory \\
\textbf{Parking\_02}    & 2270  & 88.06  & 76  & Loop in an indoor parking area. & Indoor, Short trajectory \\
\textbf{Stairs\_01}     & 9394  & 298.42 & 313 & Ascending from 2F to 6F, then looping on 6F. & Indoor, Vertical motion, Repetitive view, Long trajectory \\
\textbf{Stairs\_02}     & 4143  & 122.23 & 139 & Repeated ascending and descending of stairs. & Indoor, Vertical motion, Repetitive view, Medium trajectory \\
\textbf{Station\_01}    & 1715  & 170.84 & 229 & Escalator traversal inside a train station. & Indoor/Outdoor, Vertical motion, Repetitive view, Medium trajectory \\
\midrule
\end{tabular}
}
\begin{tablenotes}[flushleft]
\item \footnotesize Trajectory length tags : 0--100\,m = Short;\; 100--200\,m = Medium;\; $>$ 200\,m = Long.
\end{tablenotes}
\vspace{-3mm}
\end{threeparttable}
\label{dataset_summary}
\end{table*}
\begin{figure*}[t]
    \centering
    \includegraphics[width=0.9\textwidth]{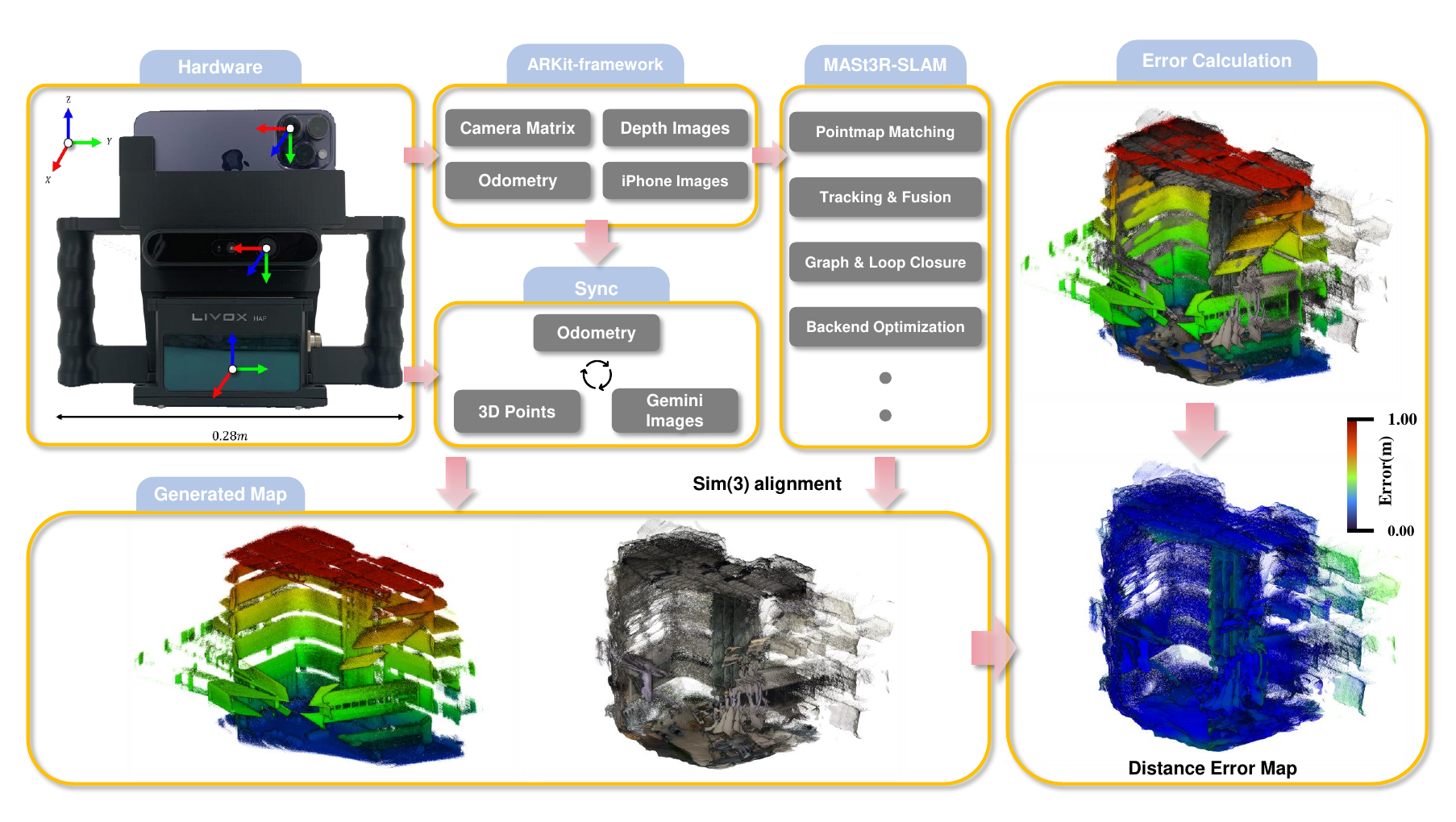}
    \caption{Our overall experimental pipeline, illustrating the process from data acquisition with our custom rig (left), through ground truth map generation and SLAM processing (center), to the final map-to-map error calculation (right).}

    \vspace{-3mm}
    
    \label{fig:system_overview}
\end{figure*}

In conclusion, this research critically addresses the challenges of intra-session scale drift and inter-session scale ambiguity, which are rooted in the fundamental scale ambiguity of monocular visual SLAM systems.

\section{The ScaleMaster Dataset}
\label{sec:dataset}

To address the inadequacy of existing benchmarks for evaluating robustness for scale consistency in monocular visual \ac{SLAM}, we introduce and publicly release a new challenging dataset, \textbf{ScaleMaster Dataset}.

\subsection{Hardware}
\label{sec:dataset_hw}

For data acquisition, we built a custom handheld rig (see \figref{fig:system_overview}) equipped with an iPhone 14 Pro, a Livox HAP LiDAR sensor, and an Orbbec Gemini 335L camera. This setup ensures precise temporal synchronization between visual frames and LiDAR scans, leading to improved data quality and reliability. The iPhone provides ARKit odometry, while the LiDAR sensor captures dense 3D geometry. This hardware configuration enables the collection of long, large-scale trajectories with consistent temporal alignment.

\subsection{Baseline Pose Generation}
\label{sec:dataset_baseline}

The baseline camera trajectories were obtained using Apple’s ARKit framework, which is known to provide centimeter-level accuracy in large indoor spaces. Since our primary goal is to highlight the fundamental scale failures, ARKit trajectories that may potentially exhibit long-term drift are sufficient, where residual centimeter-scale errors are tolerable. To build reference maps, we projected high-resolution LiDAR point clouds onto the ARKit trajectories. As shown in \figref{fig:system_overview}, this process yields dense maps that are qualitatively well aligned and preserve fine structural details, making them a reliable reference for analyzing scale consistency and geometric distortion.

\subsection{Dataset Summary and Comparison}
\label{sec:dataset_summary}

The ScaleMaster dataset comprises 25 sequences that specifically address scale-related challenges rarely captured in previous benchmarks. Of these, 18 sequences, acquired solely with an iPhone 14 Pro, provide trajectory-only ground truth, while the remaining 7 sequences, collected using the rig shown in \figref{fig:system_overview}, also include dense LiDAR-based 3D maps. The dataset covers multi-floor motion, long loops with repetitive structures, large open spaces, and low-texture conditions.

\begin{itemize}
  \item \textbf{\tabref{dataset_comparison}} compares ScaleMaster against standard benchmarks, highlighting its unique support for scale evaluation in complex indoor environments.

  \item \textbf{\tabref{dataset_summary}} presents representative sequences with their scale, duration, and environmental tags, illustrating the dataset’s diversity—from nearly 900 m building-scale loops to vertical stair traversals.

\end{itemize}

Each sequence is named by its location followed by an index number (e.g., Library\_01, Parking\_02). The environment description and trajectory characteristics for each sequence are summarized as tags in \tabref{dataset_summary}, covering attributes such as trajectory type, vertical motion, repetitive views, and low-texture conditions. These characteristics establish ScaleMaster as a challenging benchmark for testing SLAM robustness to intra- and inter-session scale inconsistency.

\section{Benchmark Evaluation}
\label{sec:exp}

In this section, we present our comprehensive evaluation of state-of-the-art monocular deep visual SLAM systems, designed to test their limits on ScaleMaster benchmark. Our complete experimental workflow—from data acquisition with our custom rig, through metric SE(3) pose estimation and baseline metric map generation, to the final map-to-map error calculation—is illustrated in \figref{fig:system_overview}. 

Our analysis proceeds in three stages: we begin by applying conventional trajectory metrics to uncover Sim(3) pose estimation failure cases, then demonstrate the inherent limitations of relying on these metrics alone, and finally introduce our direct map-to-map quality evaluation \cite{wu2021density}, which provides evidence of geometric inconsistencies further supported by qualitative visualizations.

\subsection{Experimental Setup}
\label{sec:exp-setup}

\begin{itemize}
  \item \textbf{Baselines:} 
  We evaluate three representative deep learning-based SLAM systems: \text{DROID-SLAM} \cite{teed2021droid}, \text{MASt3R-SLAM} \cite{murai2025mast3r}, and \text{VGGT-SLAM} \cite{maggio2025vggt}. All experiments were conducted using the official implementations provided by the authors, on a desktop with an AMD Ryzen 9 9900X CPU and an NVIDIA RTX 5090 GPU.
  
  \item \textbf{(Pose-oriented) Trajectory Evaluation:} 
  All trajectories are evaluated using the \ac{ATE} in \ac{RMSE} (m) using the \texttt{evo} evaluator \cite{grupp2017evo}. To ensure comparability, a $\text{Sim}(3)$ alignment \cite{umeyama2002least} was applied for scale correction because our primary focus is on intra- and inter-scale consistency challenges, rather than on metric scale evaluation (i.e., a single global scale parameter estimation).
  
  \item \textbf{(Map-oriented) 3D Reconstruction Quality Evaluation:} 
  We assess geometric fidelity by aligning the SLAM-generated map to a LiDAR point cloud, which we treat as a near–ground-truth metric map. We apply the scale, rotation, and translation $(s,R,t)$ parameters derived from the aforementioned \texttt{evo} trajectory alignment based on Umeyama algorithm. This method tests the hypothesis that a correct trajectory should yield a correct map under the same transformation. After alignment, we compute the Chamfer distance and the Drop Rate(\%); we define the Drop Rate as the percentage of points in the SLAM-generated map that have no corresponding point in the ground truth metric map within a predefined distance threshold. This metric is designed to quantify severe outliers and regions of complete map failure.
\end{itemize}

\subsection{Trajectory Evaluation}
\label{sec:exp-quant}

We begin our quantitative analysis with the standard \ac{ATE} to first establish a performance baseline and then reveal the critical failure modes that our benchmark uniquely exposes. As shown in  \tabref{arkit}, leading algorithms like MASt3R-SLAM reconfirm their state-of-the-art performance on the ARKitScenes dataset \cite{dehghan2021arkitscenes}, which establishes a baseline of their high performance under controlled, small-sized room conditions. This result can overstate general robustness.
However, this perception of robustness does not hold when evaluated on the targeted challenges of our ScaleMaster dataset, as detailed in \tabref{tab:benchmark_our}. For instance, in long-range sequences like \texttt{LargeHall\_01}, the trajectory error surges to the 80-90 meter range. This is not a random tracking loss but a direct consequence of accumulated scale drift over a long trajectory. A similar phenomenon can be observed in a different sequence, as illustrated in \figref{fig:long_term_7}.

These contrasting ATE results serve as a clear numerical demonstration that an algorithm's stability can be severely compromised by the scale-related challenges present in real-world environments, a weakness that remains hidden when evaluated on simpler, less complex benchmarks.

\begin{table}[t]
\centering
\caption{Absolute Trajectory Error (ATE (m)) on ARKitScenes sequences \cite{dehghan2021arkitscenes}.}
\begin{threeparttable}
\resizebox{0.75\columnwidth}{!}{%
\begin{tabular}{l|cccc}
\midrule
{\small \textbf{Sequence}} & 
\makecell{\textbf{DROID}\\\textbf{SLAM}} & 
\makecell{\textbf{VGGT}\\\textbf{SLAM}} & 
\makecell{\textbf{MASt3R}\\\textbf{SLAM}} & 
\makecell{\textbf{MASt3R}\\\textbf{SLAM*}} \\ 
\midrule
40777073  & 0.65 & 0.24 & 0.18 & \textbf{0.13} \\
40958754  & 0.20 & 0.03 & 0.09 & \textbf{0.02} \\
40958756  & 0.21 & 0.07 & 0.08 & \textbf{0.03} \\
41007589  & 0.32 & 0.05 & 0.09 & \textbf{0.03} \\
41045408  & \textbf{0.01} & 0.06 & 0.05 & \textbf{0.01} \\
41048083  & 0.77 & 0.82 & \textbf{0.08} & \textbf{0.08} \\
41048120  & 0.06 & 0.39 & 0.06 & \textbf{0.05} \\
\midrule
\textbf{Average} & 0.32 & 0.24 & 0.09 & \textbf{0.05} \\
\midrule
\end{tabular}}

\begin{tablenotes}[flush]
\item  \footnotesize * : calibrated mode
\end{tablenotes}
\end{threeparttable}


\label{arkit}
\end{table}
\begin{table}[t!]
\centering
\caption{Absolute Trajectory Error (unit: meter) on our dataset.}
\begin{threeparttable}
\resizebox{0.95\columnwidth}{!}{%
\begin{tabular}{l|cccc}
\midrule
{\small \textbf{Sequence}} & 
\makecell{\textbf{DROID}\\\textbf{SLAM}} & 
\makecell{\textbf{VGGT}\\\textbf{SLAM}} & 
\makecell{\textbf{MASt3R}\\\textbf{SLAM}} & 
\makecell{\textbf{MASt3R}\\\textbf{SLAM*}} \\ 
\midrule
{\footnotesize Basement\_01} &\textbf{0.08}& 1.44 & 0.38 & 0.42 \\
{\footnotesize HotelRoom\_01} & 0.05 & -- & 0.10 & \textbf{0.06} \\
{\footnotesize Lab\_01} & 0.36 & -- & 0.36 & \textbf{0.09} \\
{\footnotesize LargeHall\_01} & 89.35 & -- & \textbf{80.54} & 91.62 \\
{\footnotesize LargeHall\_02} & \textbf{3.78} & 21.69 & 6.12 & 5.89 \\
{\footnotesize LargeHall\_03} & 13.21 & -- & 1.99 & \textbf{1.96} \\
{\footnotesize LargeHall\_04} & 4.01 & 1.12 & \textbf{0.57} & 0.92 \\
{\footnotesize LargeHall\_05} & 0.56 & 0.51 & 0.45 & \textbf{0.33} \\
{\footnotesize Library\_01} & \textbf{1.68} & -- & 5.29 & 3.61 \\
{\footnotesize Library\_02} & 1.45 & -- & 0.54 & \textbf{0.63} \\
{\footnotesize Library\_03} & 0.09 & -- & 0.09 & \textbf{0.06} \\
{\footnotesize Library\_04} & 4.86 & -- & 3.54 & \textbf{3.22} \\
{\footnotesize Library\_05} & 4.35 & 13.26 & \textbf{3.08} & 4.00 \\
{\footnotesize Library\_06} & 0.05 & -- & 0.05 & \textbf{0.04} \\
{\footnotesize Library\_07} & 0.13 & 0.22 & 0.13 & \textbf{0.12} \\
{\footnotesize Library\_08} & 0.09 & -- & 0.09 & \textbf{0.06} \\
{\footnotesize Library\_09} & 0.04 & -- & 0.07 & \textbf{0.05} \\
{\footnotesize Lobby\_01} & 0.76 & 3.18 & 0.54 & \textbf{0.27} \\
{\footnotesize Lounge\_01} & 4.51 & -- & 0.47 & \textbf{0.16} \\
{\footnotesize Office\_01} & 5.61 & -- & 8.03 & \textbf{0.65} \\
{\footnotesize Parking\_01} & \textbf{10.21} & -- & 32.37 & 26.13 \\
{\footnotesize Parking\_02} & \textbf{0.20} & -- & 0.39 & 0.21 \\
{\footnotesize Stairs\_01} & 20.20 & -- & 4.60 & \textbf{2.30} \\
{\footnotesize Stairs\_02} & 5.59 & 1.05 & 1.00 & \textbf{0.14} \\
{\footnotesize Station\_01} & 11.66 & -- & 13.21 & \textbf{4.37} \\
\midrule
\end{tabular}%
} 
\vspace{2mm}
\begin{tablenotes}[flushleft]
\item \footnotesize
\parbox{0.98\columnwidth}{%
* : calibrated mode \\
-- : Runs that terminated with an invalid pose update (negative determinant during SL(4) normalization) were marked as failures; ATE is therefore undefined for those sequences.
}
\end{tablenotes}
\end{threeparttable}
\label{tab:benchmark_our}
\end{table}

\begin{table}[t]
\centering
\caption{3D Reconstruction Quality of MASt3R-SLAM}
\label{table:recon_quality}
\resizebox{1.0\columnwidth}{!}{%
\begin{tabular}{l|cccc}
\hline
\toprule
\makecell{\textbf{Sequence}\\\textbf{Name}} & 
\makecell{\textbf{Threshold} } & 
\makecell{\textbf{Chamfer}\\\textbf{Distance(m)}} & 
\makecell{\textbf{Drop}\\ \textbf{Rate}(\%)} & 
\textbf{Analysis} \\
\midrule
\multirow{2}{*}{\makecell[l]{\textbf{Library\_01}}}
  & 1  & 0.36 & 42.5 & \textbf{Severe Failure} \\
  & 10 & \textbf{6.43} & 0.9 & Significant map distortion. \\
\hline
\multirow{2}{*}{\textbf{Library\_06}} 
  & 1  & 0.08 & 1.1 & \textbf{Success} \\
  & 10 & \textbf{0.10} & 0.0 & High-fidelity reconstruction. \\
\hline
\multirow{2}{*}{\textbf{Library\_07}} 
  & 1  & 0.52 & \textbf{89.1} & \textbf{Catastrophic Failure} \\
  & 10 & \textbf{9.99} & 0.0 & Near-total map collapse. \\
  
\bottomrule
\hline
\end{tabular}}
\label{map_qual}
\end{table}
\begin{figure}[!t]
    \centering
    \includegraphics[width=1.0\columnwidth]{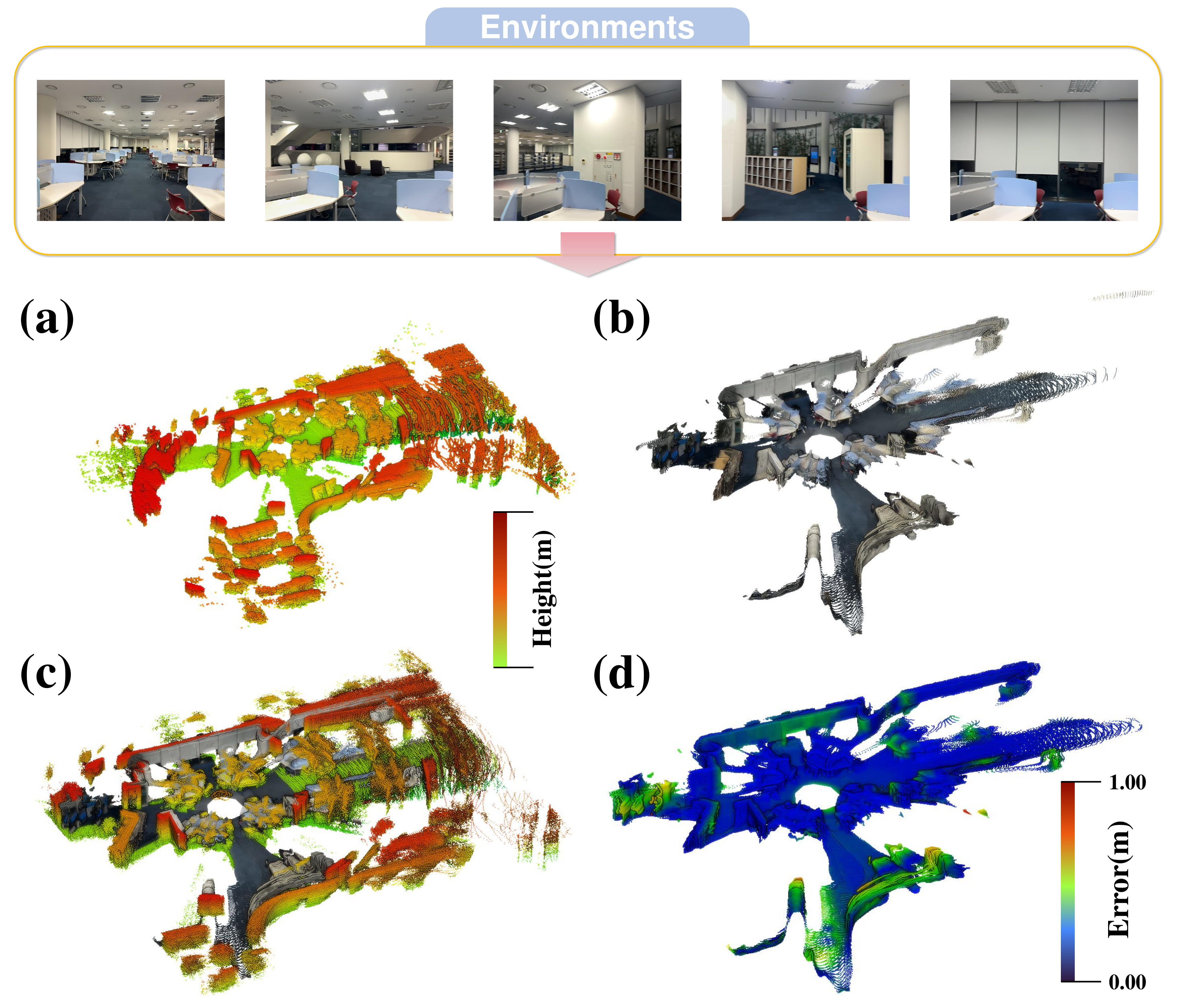}
    \caption{Qualitative comparison of 3D reconstruction of the \texttt{Library\_06} sequence. (a) Ground truth point cloud from LiDAR, color-coded by height. (b) The map reconstructed by MASt3R-SLAM. (c) The alignment of the MASt3R-SLAM map onto the ground truth. (d) Point-to-point distance error visualization, where warmer colors (red) indicate larger geometric inconsistencies between the two maps}
    
    \vspace{-3mm}
    
    \label{fig:lib6_fig}
\end{figure}
\begin{figure}[t]
    \centering
    \includegraphics[width=1.0\columnwidth,height=0.3\textheight]{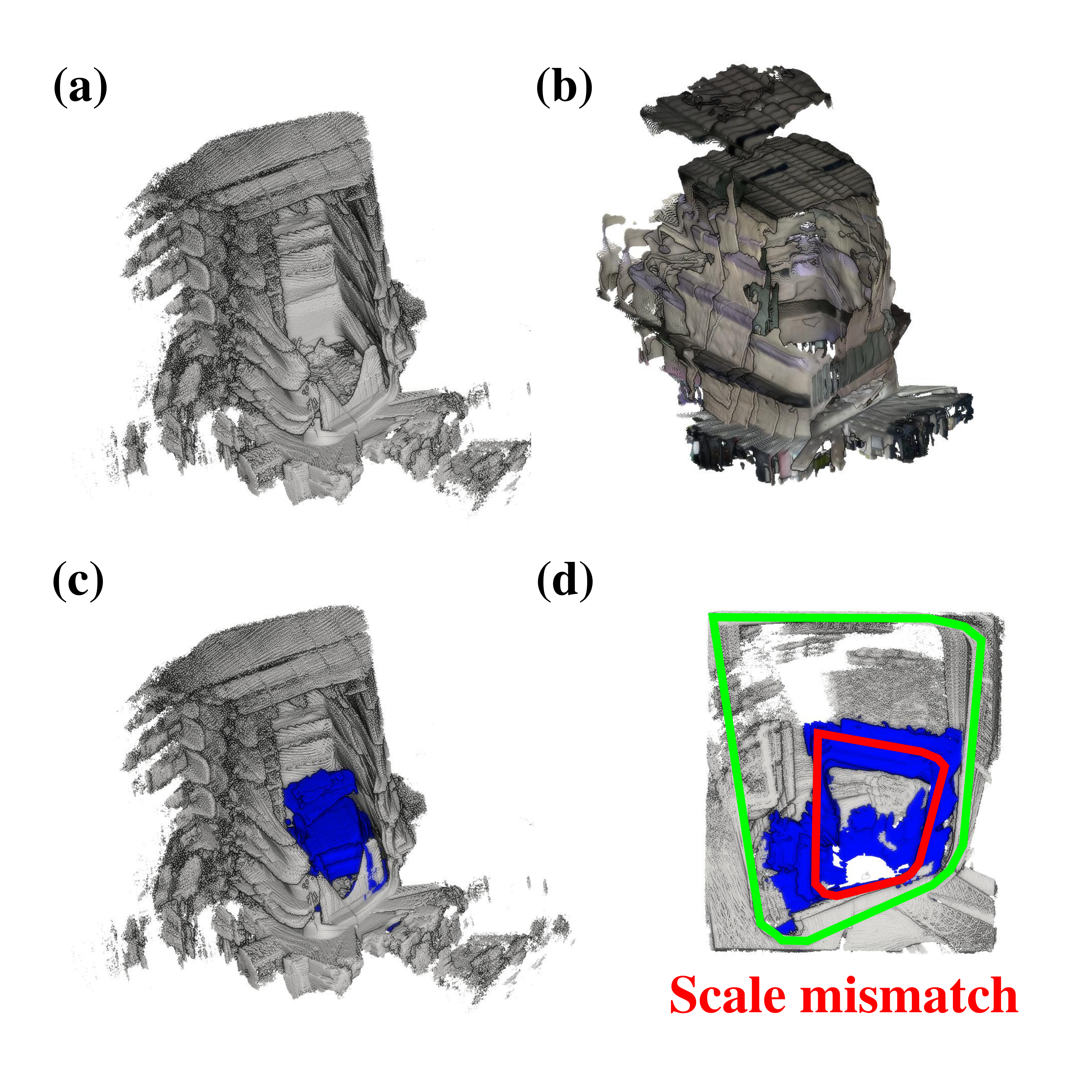}
    \caption{Qualitative comparison on \texttt{Library\_07} demonstrating that low trajectory error does not guarantee correct map scale. (a) The LiDAR ground-truth and (b) the MASt3R-SLAM reconstruction. (c) The overlay and (d) top-down cross-section are generated after applying the single global Sim(3) transformation calculated to align the camera trajectories.}

    \vspace{-3mm}
    
    \label{fig:lib1_fig}
\end{figure}
\subsection{3D Reconstruction Quality: Exposing Geometric Failures}
\label{sec:exp_map}

While the dramatic \ac{ATE} failures are informative, \ac{ATE} provides only a partial characterization of system performance. It does not describe the geometric fidelity and completeness of a resulting dense 3D map, which is the primary output of a dense visual SLAM system. As our subsequent analysis will prove, a SLAM system can produce a trajectory with low ATE, yet the corresponding map can be severely warped or incorrectly scaled due to internal scale inconsistencies. Therefore, a more direct evaluation of the 3D map is essential \cite{hu2025mapeval}.

To overcome the limitations of ATE, we use Chamfer distance and Drop Rate mentioned in \secref{sec:exp-setup}. These direct map-to-map quality evaluations are presented in \tabref{map_qual}. This analysis reveals the metric reconstruction failures.

\begin{itemize}
  \item \textbf{Success Case}:
  On the \texttt{Library\_06} sequence (see \figref{fig:lib6_fig}), the baseline algorithm (e.g., MASt3R-SLAM) performs well in both pose estimation (0.04 m in \tabref{tab:benchmark_our}) and map reconstruction up to a single global scale, achieving a low Chamfer distance of 0.10 m. This example is the case where the scale is well maintained, thus both the trajectory and the map are correct.

  \item \textbf{Catastrophic Failure Case}: 
  In contrast, \figref{fig:lib1_fig} shows that the \texttt{Library\_07} sequence suffers a geometric failure even though the pose error is low (0.12 m in \tabref{tab:benchmark_our}). With the correspondence distance threshold set to 1 m, 89.1\% of the generated map points were discarded as outliers, and when the threshold was increased to 10 m, the Chamfer distance reached a massive 9.99 m. This scale consistency-aware reconstruction failure is fundamentally invisible to ATE but is captured perfectly by the map quality metrics.  
\end{itemize}
These results show that trajectory error alone, even after a single scalar scale adjustment, is insufficient and often misleading. Direct map quality evaluation is therefore necessary to assess dense visual SLAM reliably.

\subsection{Qualitative Analysis of Scale Inconsistency}
\label{sec:exp_qual}

\begin{figure}[t]
    \centering
    \includegraphics[width=0.95\columnwidth]{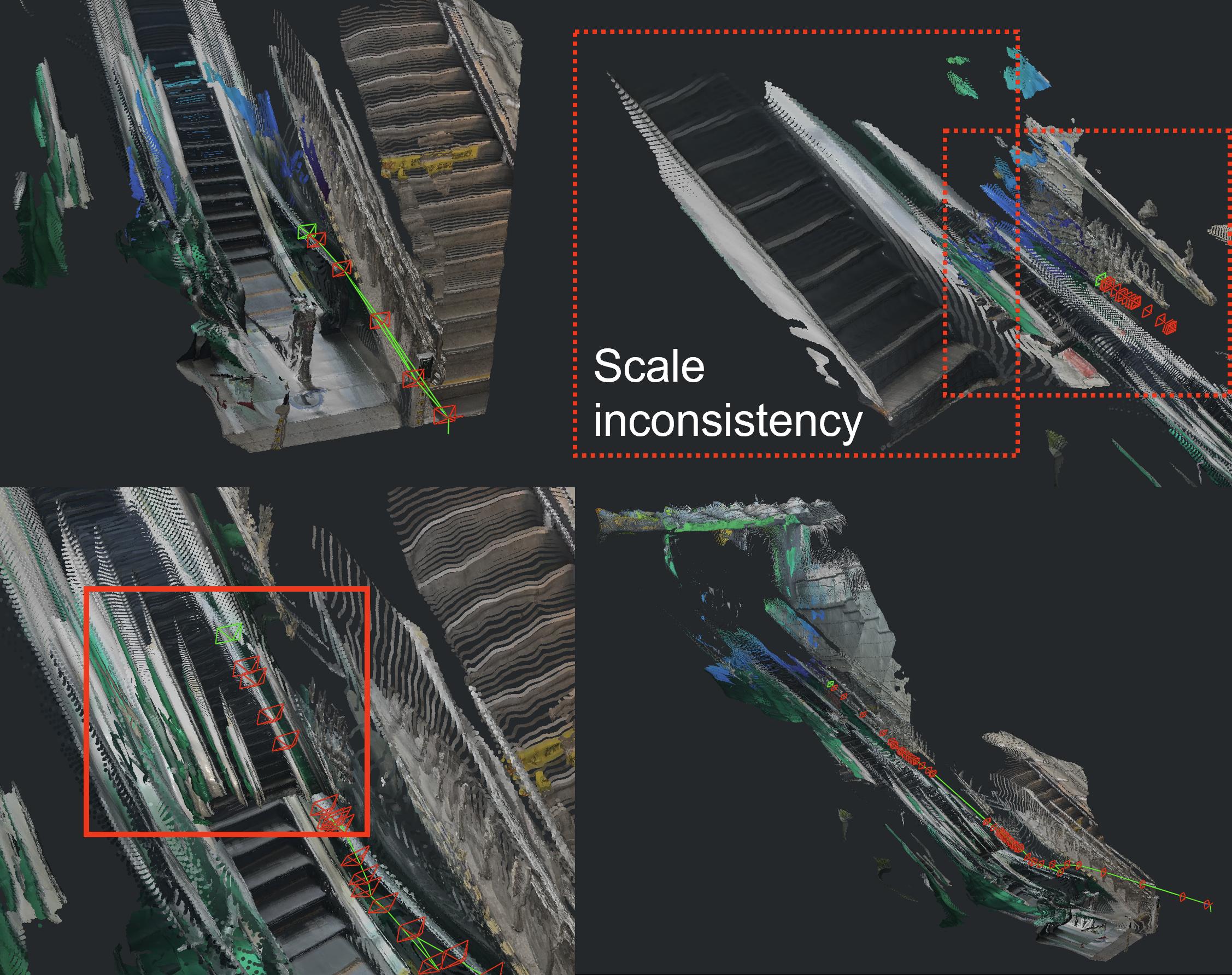}
    \caption{Short-term scale inconsistency of MASt3R-SLAM on the \texttt{Station\_01} sequence. The repetitive structure and vertical motion cause a severely distorted and overlapping 3D map. This demonstrates the system's vulnerability even in short, challenging trajectories.}


    \label{fig:short_term_scale_inconsistency}
\end{figure}

\begin{figure}[t!]
    \centering
      \begin{overpic}[width=0.95\linewidth]{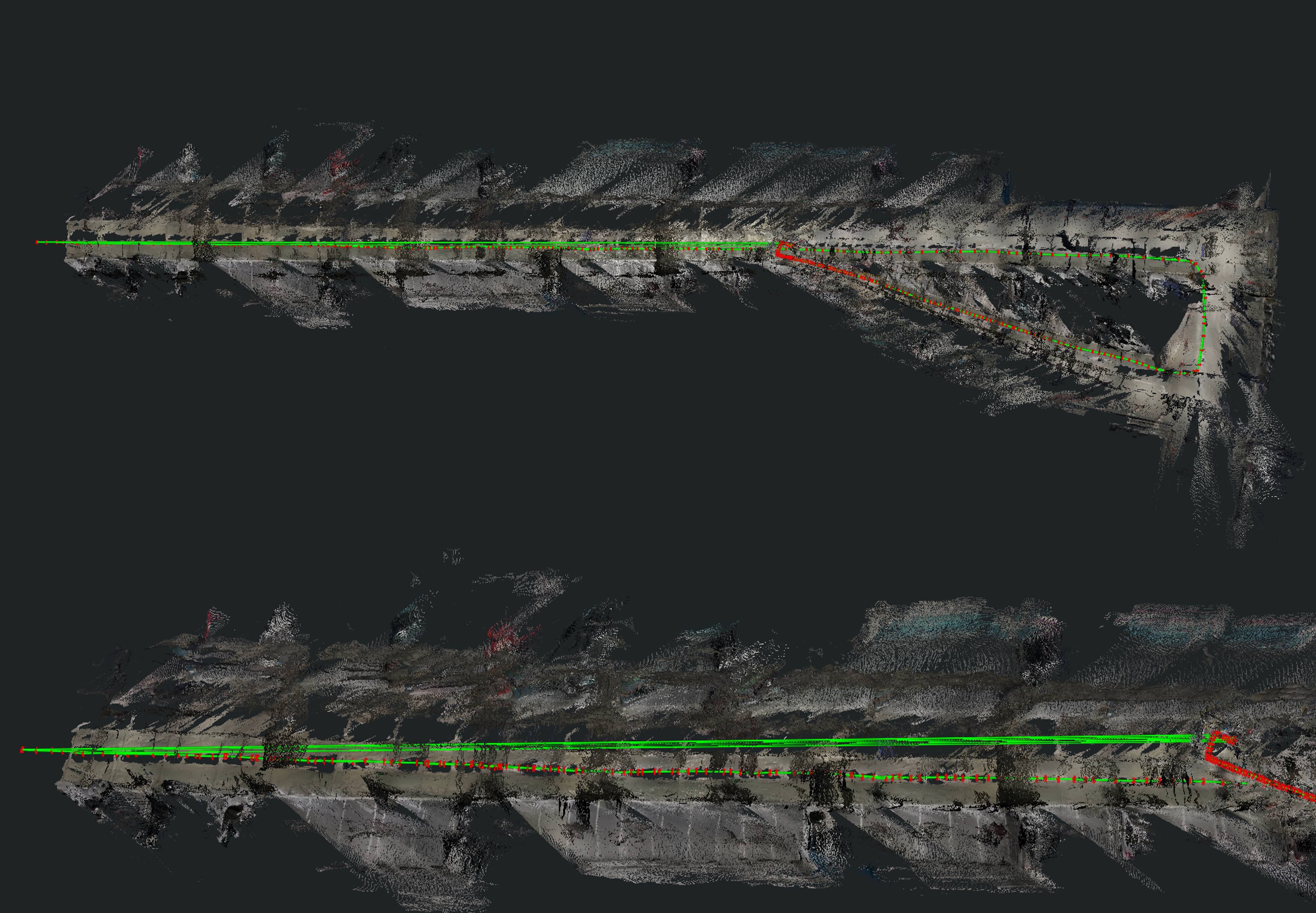}
        \put(1,66){\color{white}\scriptsize Actual trajectory returns to origin but Sim(3) drift accumulates}
        \put(1,30){\color{white}\scriptsize Loop closure detected but Sim(3) optimization fails}
      \end{overpic}
        
    \caption{A long-term optimization failure case for MASt3R-SLAM on the \texttt{Parking\_01} sequence.}


    \label{fig:long_term_7}

\end{figure}

\begin{figure}[!t]
    \centering
    \includegraphics[width=1.0\columnwidth]{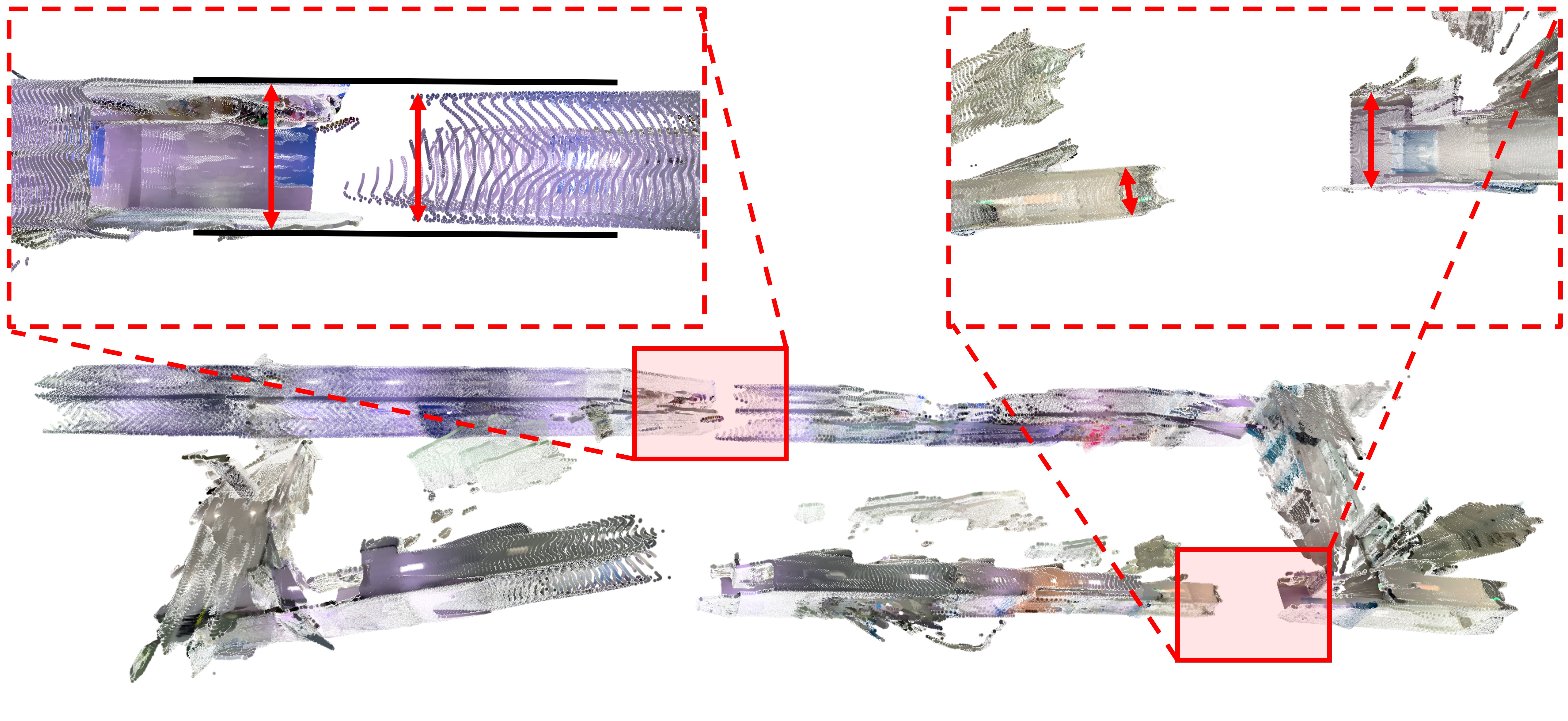}
    \caption{A failure case of inter-session scale ambiguity in MASt3R-SLAM. The image shows the result of processing a single sequence as three independent sessions. The resulting map fragments cannot be aligned into a consistent map as each was generated with a different, inconsistent scale.}

    \vspace{-3mm}

    \label{fig:multi_scale_inconsistency}
\end{figure}

Building on the taxonomy of scale inconsistency outlined in \figref{fig:problem_diagram}, we now illustrate three representative failure cases, each corresponding to a distinct category of the problem. The following three figures (\figref{fig:short_term_scale_inconsistency}, \figref{fig:long_term_7}, and \figref{fig:multi_scale_inconsistency})  serve as visual counterparts to the taxonomy in Fig. 2, grounding the abstract categories in concrete empirical evidence.

\noindent \textbf{Intra-session Scale Inconsistency:} \figref{fig:short_term_scale_inconsistency} illustrates short-term scale failure in a vertical motion sequence, \texttt{Station\_01}. Repetitive stair patterns disrupt data association between consecutive frames, causing immediate scale drift and severe map distortions like multi-layer overlaps. Furthermore, \figref{fig:long_term_7} shows a long-term optimization failure on the \texttt{Parking\_01} sequence. Although a loop closure is detected, the massive accumulated scale drift traps the optimizer in a local minimum, resulting in a geometrically inconsistent map.

\noindent \textbf{Inter-session Scale Ambiguity:} \figref{fig:multi_scale_inconsistency} demonstrates another critical issue. When a single video is processed as three independent sessions, each resulting map fragment is generated at a different, inconsistent scale. While internally coherent, they cannot be merged into a single, globally scale-consistent map, highlighting a major challenge for long-term mapping or collaborative SLAM.   

%

%


%
\section{Conclusion}
\label{sec:conclusion}

In this work, we examined the fundamental challenges of three types of scale inconsistencies in deep monocular visual \ac{SLAM}. With ScaleMaster, which exposes intra-session scale drift and inter-session scale ambiguity in large indoor environments, we coupled traditional trajectory evaluation with a direct map-to-map assessment (e.g., Chamfer distance and Drop Rate). Across multiple baselines (including DROID-SLAM, MASt3R-SLAM, and VGGT-SLAM), our quantitative and qualitative results reveal severe scale-related failures under ScaleMaster's more realistic conditions, despite good performance on existing benchmarks. We hope this dataset and the accompanying baseline evaluations provide a solid platform for measuring and improving scale-consistent, reliable SLAM.

\section*{ACKNOWLEDGMENT}
This work was supported by Basic Science Research Program through the National Research Foundation of Korea (NRF) funded by the Ministry of Education (No.RS-2025-25420118), Institute of Information \& Communications Technology Planning \& Evaluation (IITP) grant funded by the Korea government (MSIT) (No.RS-2025-02219277, AI Star Fellowship Support (DGIST)), and the InnoCORE program of the Ministry of Science and ICT (26-InnoCORE-01).




\bibliographystyle{unsrt}
{\small
\bibliography{ref}
}

\end{document}